# RCdpia: A Renal Carcinoma Digital Pathology Image Annotation dataset based on pathologists

Qingrong Sun[1], Weixiang Zhong[2], Jie Zhou[2], Chong Lai[3], Xiaodong Teng[2,*] and Maode Lai[4,*]

[1]College of Information Technology, ZheJiang Shuren University, Hangzhou, 310000, Hangzhou, China.

[2]Department of Pathology, The First Affiliated Hospital, School of Medicine, Zhejiang University, Hangzhou, 310003, China

[3]Department of Urology, The First Affiliated Hospital, School of Medicine, Zhejiang University, Hangzhou, 310003, China

[4]Department of Pathology, Research Unit of Intelligence Classification of Tumor Pathology and Precision Therapy, Chinese Academy of Medical Science, Alibaba-Zhejiang University Joint Research Center of Future Digital Healthcare, Key Laboratory of Disease Proteomics of Zhejiang Province, Zhejiang University School of Medicine, Hangzhou 310058, China

*Correspondence: teng1723@163.com (Xiaodong Teng) and lmd@zju.edu.cn (Maode Lai)

**Abstract**

The annotation of digital pathological slide data for renal cell carcinoma is of paramount importance for correct diagnosis of artificial intelligence models due to the heterogeneous nature of the tumor. This process not only facilitates a deeper understanding of renal cell cancer heterogeneity but also aims to minimize noise in the data for more accurate studies. To enhance the applicability of the data, two pathologists were enlisted to meticulously curate, screen, and label a kidney cancer pathology image dataset from The Cancer Genome Atlas Program (TCGA) database. Subsequently, a Resnet model was developed to validate the annotated dataset against an additional dataset from the First Affiliated Hospital of Zhejiang University. Based on these results, we have meticulously compiled the TCGA digital pathological dataset with independent labeling of tumor regions and adjacent areas (RCdpia), which includes 109 cases of kidney chromophobe cell carcinoma, 486 cases of kidney clear cell carcinoma, and 292 cases of kidney papillary cell carcinoma. This dataset is now publicly accessible at http://39.171.241.18:8888/RCdpia/. Furthermore, model analysis has revealed significant discrepancies in predictive outcomes when applying the same model to datasets from different centers. Leveraging the RCdpia, we can now develop more precise digital pathology artificial intelligence models for tasks such as normalization, classification, and segmentation. These advancements underscore the potential for more nuanced and accurate AI applications in the field of digital pathology.

**Introduction**
 Telepathology, the pioneering digital branch of pathology, was first introduced in an academic lecture by Dr. Weinstein in 1986[1]. Since its inception, the practice of remote pathological diagnosis has gained global recognition, prompting some countries to establish dedicated networks to facilitate these services [2-4]. The term 'digital pathology' has evolved over time, transitioning from the concept of a virtual microscope [5] to the advent of whole slide image (WSI) [6]. The turn of the millennium marked a rapid acceleration in the development of digital pathology, with a corresponding surge in related research. An analysis of studies from 1991 to 2022, using "digital pathology" as a keyword, corroborates this upward trajectory [7-9]. The critical role of WSI normalization in model performance has come to light through extensive research, leading to ongoing studies in areas such as WSI color standardization based on artificial intelligence model [10] and pathologist-score-based WSI standardization [11]. Pathological diagnosis, the "gold standard" of medical diagnosis, underpins the implementation of nearly all therapeutic interventions, which are contingent upon accurate pathological diagnosis. Similarly, personalized tumor therapy, or precision medicine relies on the precise molecular classification of tumor pathology. Intelligent pathological classification is thus fundamental to the precision diagnosis and treatment of cancer patients. Without accurate subtyping, individualized treatment regimens would be unattainable.
Renal cell carcinoma, comprising various tissue types such as kidney clear cell carcinoma (KIRC), kidney chromophobe cell carcinoma (KICH), and kidney papillary cell carcinoma (KIRP) presents a diverse morphological character. KIRC, the most subtype form of renal cell cancer, constitutes approximately 70%-75% of the all renal cancers [12]. Over the past two decades, the incidence of kidney cancer has surged by an annual rate of 2%, representing about 3% of all malignant tumors [13].
 In light of these statistics, this study sourced digital pathological data of patients with renal cancer from The Cancer Genome Atlas Program (TCGA) database [14] to construct a renal cancer annotation dataset. Furthermore, two ResNet classification models were developed to assess the reliability of the renal cancer classification, these models were subsequently validated by the First Affiliated Hospital, Zhejiang University school of Medicine. The ultimate goal is to enable the models to accurately delineate each region (tumor or normal) of renal cell cancer WSI.

**Methods**
**Sample preparation**
 We procured a total of 928 digital pathology images of renal cell cancer from the TCGA database, encompassing 519 cases of Kidney Renal Clear Cell Carcinoma (KIRC), 109 cases of Kidney Chromophobe (KICH), and 300 cases of Kidney Renal Papillary Cell Carcinoma (KIRP). Additionally, we acquired Whole Slide Images (WSIs) of renal cell cancer cases, including 142 KICH, 373 KIRC, and 167 KIRP, from the First Affiliated Hospital of Zhejiang University School of Medicine (FAHZU). These were utilized to evaluate the accuracy of our classification model, which was informed by our meticulously annotated data.
 Subsequently, we undertook a comprehensive process of filtering, annotating, and segmenting all WSIs. Leveraging the dataset, we categorized the labels into comprehensive tumor outlines (tALL), characteristic tumor regions (tumor), and adjacent normal tissues (normal). Utilizing this stratification, we developed ResNet-18 and ResNet-50 models to accurately predict the subtypes of renal cell cancer (KICH, KIRC, and KIRP). The intricate details of our framework are depicted in

Figure 1.

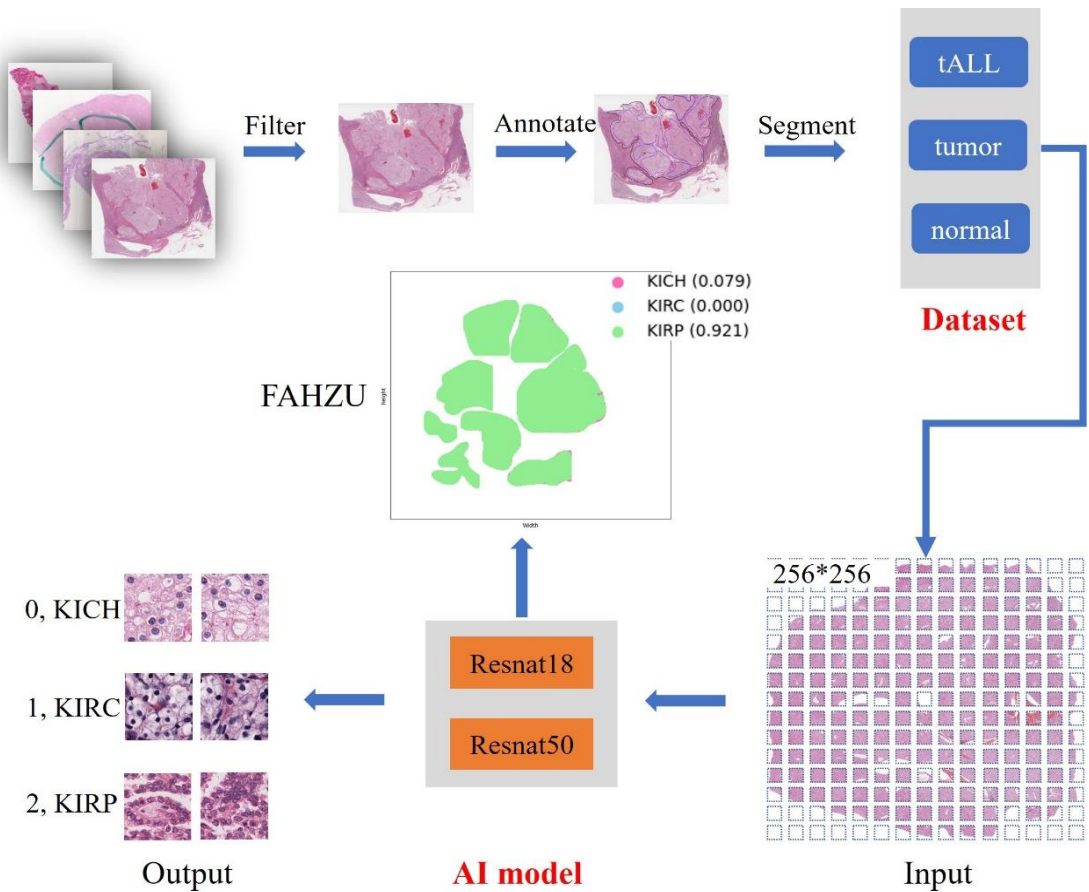

**Figure 1. Framework for RCC digital pathology annotation.** FAHZU means the First Affiliated Hospital of Zhejiang University, School of Medicine, KICH means kidney chromophobe cell carcinoma, KIRC means kidney clear cell carcinoma, and KIRP means kidney papillary cell carcinoma.

**Framework for WSI Annotation**

To facilitate precise segmentation of each cancer subtype within the ImageScope software, we designated distinct line colors to demarcate the various subtypes, as illustrated in Figure 2. Concurrently, we engaged pathologists to meticulously annotate all digital pathological images, integrating diagnostic insights to delineate three distinct regions within each WSI: the tumor area, the typical tumor area, and adjacent normal areas.

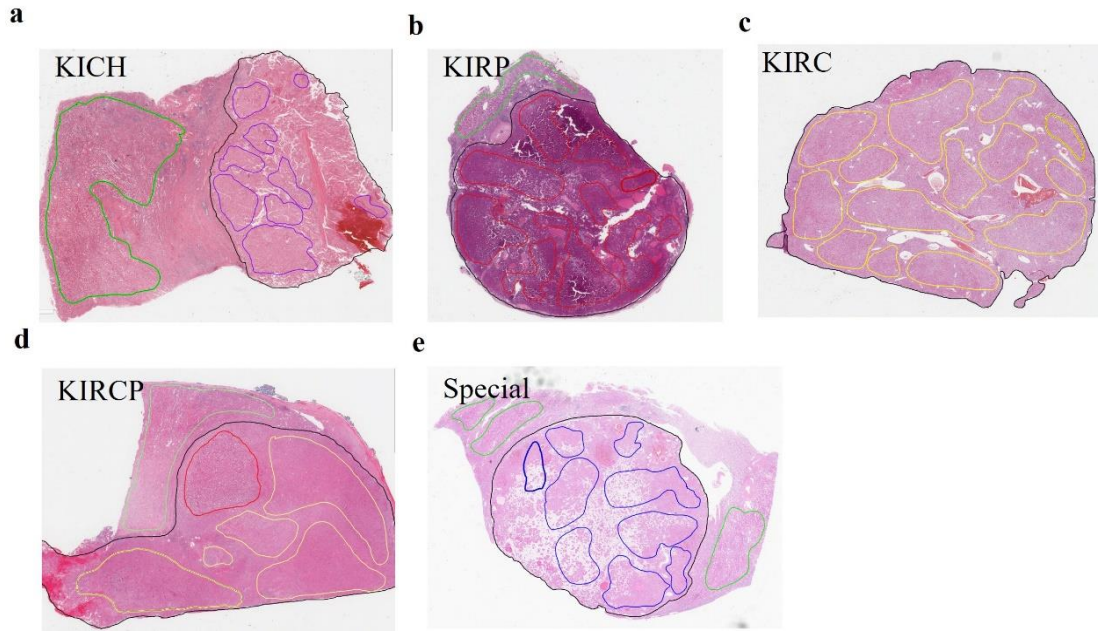

**Figure 2. The annotation color for every subtype of RCC.** a. KICH in purple color, b. KIRP in red color, c. KIRC in yellow color, d. KIRCP (means one sample have KIRC and KIRP two subtypes tumor). e. Special means other types of renal cell carcinomas excluding above subtypes from a-d with blue color.

**Clinical information matching**

We acquired additional data, including survival statistics, mRNA expression profiles, microRNA expression data, protein expression metrics, and gene somatic mutation records from the TCGA database and the UCSC Xena platform. This comprehensive collection was then synthesized to establish a kidney cancer repository, offering users direct access to annotated images, an array of omics data, and pertinent clinical prognostic information.

**Technical Validation**

For the dataset validation, we segmented all WSI of three renal cell cancer subtypes (KICH, KIRC and KIRP) into 256×256pixel patches. Utilizing these patches, we developed ResNet-18 and ResNet-50 models to facilitate the classification of renal cell cancer subtypes. The ResNet models were then validated using the FAHZU dataset. The training sets consisted of 9, 12 and 20 WSIs for each subtype from FAHZU respectively. We proceeded to evaluate the threshold for the model was set 0.5, signifying that the subtype representing more than 50% of the test results for a WSI was designated as the definitive predicted subtype for that WSIs.

**Results**

Through our labeling and screening process, we curated a collection of digital pathological images of renal cancer from the TCGA database, which included 109 cases of Kidney Chromophobe (KICH), 486 cases of Kidney Renal Clear Cell Carcinoma (KIRC), and 292 cases of Kidney Renal Papillary Cell Carcinoma (KIRP). We then segmented all annotated region of WSIs in 256×256 pix patches (Figure 2a), and quantified the number of tumor regions in KICH, KIRC and KIRP (Figure 2b-2c). The subsequent phase involved developing ResNet-18 and -50 models based on the TCGA dataset, which was partitioned into an 80% training set and a 20% testing set. These models achieved an approximate accuracy of 99% across the three subtypes, as shown in Figure 2d and 2e.

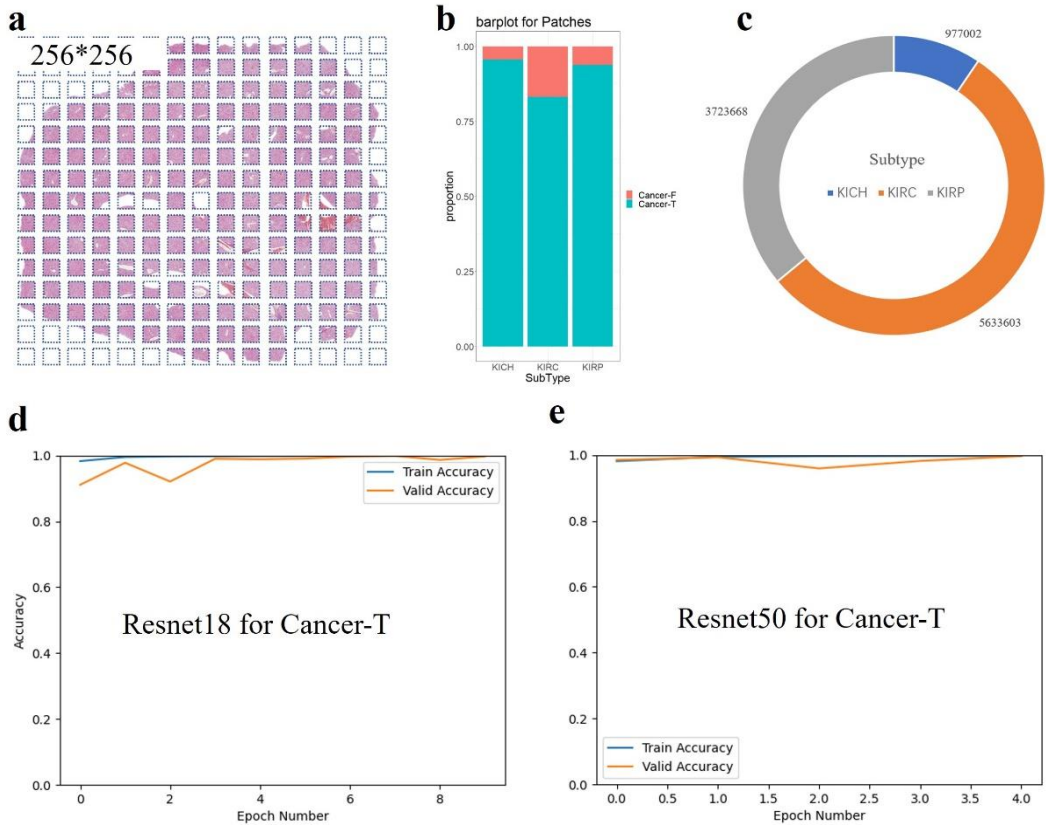

**Figure 3. 256×256 Patches and model results for RCC WSIs from the TCGA.** a represent a diagram for 256*256 patches segmentation, b shows the proportion for Cancer regions of every subtype (KICH, KIRC and KIRP), c shows the Cancer-T distribution in every subtype (KICH, KIRC and KIRP), d-e show the model accuracy for classification of RCC by Resnet-18 and Resnet-50. Cancer-F means cancer region for tumors in WSI and was noise patches; Cancer-T means cancer region for tumors in WSI and was true tumor patches.

To assess the robustness of the aforementioned models, we employed the FAHZU dataset for validation. We observed that the accuracy for the three subtypes improved in correlation with the enlargement of the training set. Moreover, we noted that ResNet-18 exhibited slightly lower accuracy than ResNet-50 for the KIRC and KIRP subtypes, whereas the inverse was true for KICH. Intriguingly, the highest average accuracy across the three subtypes was achieved with the ResNet-18 model. Detailed results are presented in Table 1.

**Table 1. Accuracy of ResNet18 and 50 models based on different training data sets.**

| model | ResNet18 | | | | ResNet50 | | | |
|---|---|---|---|---|---|---|---|---|
| Train set | 0 | 9 | 12 | **20** | 0 | 9 | 12 | **20** |
| KICH | 0.959 | 0.648 | 0.885 | **0.893** | 0.131 | 0.779 | 0.811 | **0.828** |
| KIRC | 0.668 | 0.932 | 0.815 | **0.946** | 0.080 | 0.866 | 0.923 | **0.943** |
| KIRP | 0.608 | 0.919 | 0.959 | **0.926** | 0.912 | 0.939 | 0.959 | **0.939** |

## Discussion

We downloaded digital pathological images of renal cell cancer from the TCGA database and annotated the tumor and adjacent normal areas. Subsequently, we established and made available the RCdpia at http://39.171.241.18:8888/RCdpia/ . During above process, we identified significant disparities in image quality, including issues such as graffiti on images, misclassification, poor clarity, and organizational errors. These WSIs were subsequently excluded.

Following this, we developed ResNet-18 and ResNet-50 classification models based on the curated TCGA WSIs, achieving an accuracy of approximately 99% with both models. However, when these models were directly applied to the FAHZU dataset, there was a marked decline in accuracy. To address this, we retrained the pretrained models using the FAHZU dataset and observed that accuracy improved proportionally with the expansion of the training data. This task highlighted a significant variance in model accuracy when applied to pathological images from different centers. To enhance the model's robustness, further work is required to normalize WSI data. While numerous studies have investigated normalization techniques for digital pathological images, often focusing on tissue staining processes, we believe there is room for innovation [11, 15-18].

Considering the intrinsic medical characteristics of digital pathology, we recognize that each WSI comprises elements such as the nucleus, cytoplasm, and various microenvironments or lesion regions (e.g., necrosis, calcification). We propose that leveraging this understanding to construct a deep learning algorithm capable of reconstructing the color space for these components could standardize the color space across all WSIs, thereby bolstering the robustness of our classification models. Indeed, there are numerous algorithms dedicated to the segmentation of nuclei in pathological tissues [19-23]. We are optimistic about making strides in this domain of normalization.

## Author Contributions

Q.S. and M.L. conceived the original idea and designed the study. X.T and C.L collected clinical data of FAHZU cohort. W.Z., J.Z. and X.T. contributed to annotate all WSIs and screen the FAHZU dataset. Q.S. and M.L. reviewed and contributed to the manuscript.

## Code Availability

The code resource was released in http://39.171.241.18:8888/RCdpia/.

## Acknowledgments

This work was supported by the National Natural Science Foundation of China (No. 82072811).

## Competing Interests

The authors declare that they have no competing interests.

## Additional Information

**Correspondence** and requests for FAHZU dataset should be addressed to M.L.
**TCGA dataset** annotation information is available at http://39.171.241.18:8888/RCdpia/.